\documentclass[journal]{IEEEtran}

\usepackage{xcolor,soul,framed} 

\colorlet{shadecolor}{yellow}
\usepackage[pdftex]{graphicx}
\graphicspath{{../pdf/}{../jpeg/}}
\DeclareGraphicsExtensions{.pdf,.jpeg,.png}

\usepackage[cmex10]{amsmath}
\usepackage{array}
\usepackage{mdwmath}
\usepackage{mdwtab}
\usepackage{eqparbox}
\usepackage{url}
\usepackage{booktabs}
\usepackage{makecell}
\usepackage{txfonts}
\usepackage{cite}
\usepackage{booktabs}
\usepackage{multirow}
\usepackage{bm}
\usepackage{booktabs}
\usepackage{multirow}
\usepackage{graphicx}
\usepackage{setspace}
\usepackage{threeparttable}
\usepackage{booktabs}

\newcolumntype{L}[1]{>{\raggedright\arraybackslash}p{#1}}

\begin{document}

\bstctlcite{IEEEexample:BSTcontrol}
    \title{BoostTaxo: Zero-Shot Taxonomy Induction via Boosting-Style Agentic Reasoning and Constraint-Aware Calibration}
  \author{Yancheng Ling,~\IEEEmembership{}Zhenlin Qin,~\IEEEmembership{}Leizhen Wang,~\IEEEmembership{}
      Zhenliang Ma~\IEEEmembership{}
     \\
  
   \thanks{Corresponding Author: Zhenliang Ma, email: zhema@kth.se}
  \thanks{Yancheng Ling is a postdoc of KTH Royal Institute of Technology, 11428 Sweden (e-mail: yancheng@kth.se).}%
  \thanks{Zhenlin Qin is a PHD of KTH Royal Institute of Technology, 11428 Sweden (e-mail: zhenlinq@kth.se).}%
  \thanks{Leizhen Wang is a PHD of  Monash University, Clayton, Australia, 3800 (e-mail: leizhen.wang@monash.edu).}%
  \thanks{Zhenliang Ma is an Associate Professor of KTH Royal Institute of Technology, 11428 Sweden (e-mail: zhema@kth.se).}%
   .}%

\markboth{
}{BoostTaxo: Zero-Shot Taxonomy Induction via Boosting-Style Agentic Reasoning and Constraint-Aware Calibration}

\maketitle

\begin{abstract}
Taxonomy induction is crucial for organizing concepts into explicit and interpretable semantic hierarchies. While existing methods have achieved promising results, their generalization, structural reliability, and efficiency remain limited, hindering their performance in zero-shot and large-scale scenarios. To overcome these limitations, we introduce BoostTaxo, a boosting-style LLM framework for zero-shot taxonomy induction. It takes a set of domain terms as inputs and performs parent identification in a coarse-to-fine manner, employing retrieval-augmented definition refinement, hybrid parent candidate selection, candidate rating, and structure-aware score calibration to improve taxonomy construction. Specifically, a lightweight LLM is used to efficiently filter candidate parents, while a large-scale LLM is employed to rank and score candidate parents for fine-grained parent selection. Structural features are further incorporated to calibrate candidate edge weights and enhance the reliability of the induced taxonomy. The unified BoostTaxo is evaluated on three public benchmark datasets, namely WordNet, DBLP, and SemEval-Sci, and achieves superior or comparable performance to state-of-the-art methods in zero-shot taxonomy induction. The ablation study validates the contribution of the hybrid parent candidate selection and the structure-aware score calibration to the overall performance. Further analysis investigates the impact of candidate selection size on taxonomy quality and presents representative case and failure studies, providing deeper insights into the effectiveness and limitations of the proposed framework. 

\end{abstract}

\IEEEpeerreviewmaketitle
\section{Introduction}

\IEEEPARstart {T}{axonomies} are tree-structured hierarchies that organize concepts based on hypernymy (\textit{is-a}) relations, providing an explicit and interpretable semantic structure for knowledge representation\cite{shen2018hiexpan,shen2017setexpan,liu2012automatic}. By arranging concepts from more general categories to more specific ones, taxonomies enable hierarchical reasoning, semantic generalization, and fine-grained concept organization. As a result, they have become a fundamental resource in a wide range of knowledge-intensive applications. Existing studies have shown that taxonomies play an important role in various domains, including question answering\cite{cambazoglu2021intent,yang2017efficiently}, web indexing and search\cite{yin2010building,wu2012probase}, query understanding\cite{hua2016understand}, personalized recommendation\cite{huang2019taxonomy,tan2022enhancing}, and ontology construction\cite{ghosh2023learning,ma2023kepl}. 

\begin{figure}
  \begin{center}
  \includegraphics[scale=0.6]{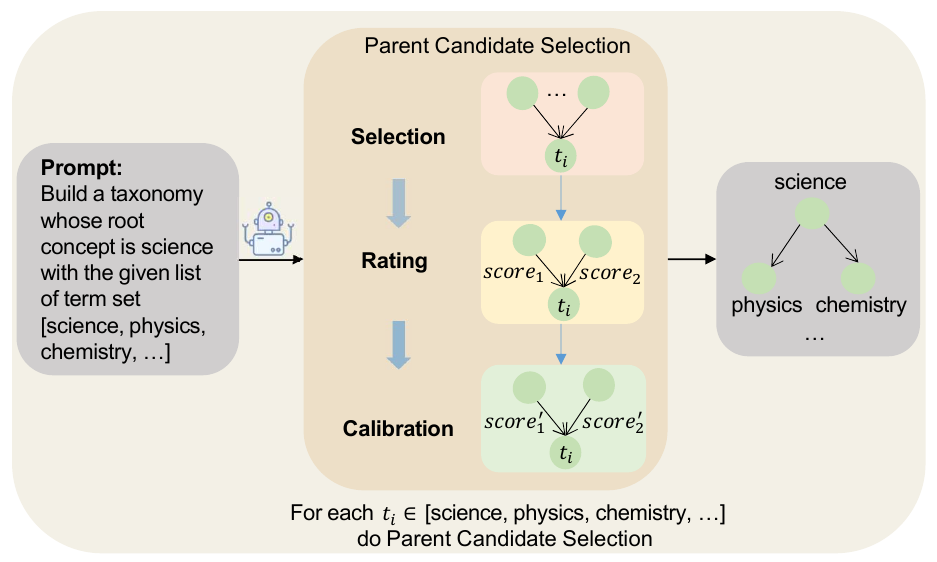}\\
 \caption{Boosting-Style Zero-Shot Taxonomy Induction with Large Language Models}\label{fig1}
  \end{center}
\end{figure}

 Traditional methods typically decompose taxonomy induction into two sequential subtasks: (1) hypernymy detection, which aims to identify term pairs exhibiting an “is-a” relation, and (2) hypernymy organization, which aims to arrange these term pairs into a tree-structured hierarchy\cite{chen2021constructing}. Of these, hypernymy detection is a crucial step in taxonomy induction. The early  methods primarily relied on identifying lexico-syntactic patterns, for example, “A such as B”, which typically indicate that B is a hyponym of A.  Although such methods are broadly applicable, their effectiveness is often limited by the sparse occurrence of indicative lexical patterns\cite{hearst1992automatic,kozareva2010semi,klaussner2011lexico,panchenko2016taxi}. To address this limitation, later studies turned to supervised and weakly supervised methods, which learn taxonomic relation classifiers or taxonomy construction models from labeled hypernym pairs or gold taxonomies and generally achieve stronger performance. However, their reliance on large amounts of domain-specific labeled data makes them difficult to generalize across domains.
 
 Recent advances in large language models (LLMs) have enabled new approaches to natural language processing tasks. Trained on massive amounts of data, LLMs possess extensive parametric knowledge. By encoding large-scale world knowledge in their parameters, they also capture latent reasoning patterns in language and exhibit strong reasoning capabilities. As a result, they have been widely applied to a variety of downstream tasks, including information extraction\cite{xu2024large,xu2023unleash}, question answering\cite{brown2020language}, text classification\cite{wang2023large,yoo2021gpt3mix}, and structured knowledge discovery\cite{ling2026review,trajanoska2023enhancing}.

LLMs have shown promising progress in taxonomy induction. Methods such as TaxonomyGPT\cite{chen2023prompting} and LM Scoring\cite{jain2022distilling} use carefully designed prompts to enable LLMs to generate hypernymy relationships among a given set of terms. Compared with traditional supervised learning approaches, these methods exhibit stronger generalization ability and have achieved encouraging results. However, they suffer from two main limitations. First, when estimating relationships and weights between terms, they often focus primarily on pairwise semantic relatedness while overlooking the structural constraints of the taxonomy. Secondly, to infer hypernymy relationships, these methods process all terms simultaneously for joint prediction. However, as the number of terms grows, such strategies can result in high computational complexity, a greater risk of token explosion, and degraded performance \cite{chen2023prompting}. To better capture structural information, Chain of Layer\cite{zeng2024chain} takes all candidate terms as input simultaneously and decomposes taxonomy induction into a layer-wise process, selecting relevant candidate entities at each level and progressively constructing the taxonomy from top to bottom. Although this method has achieved promising results, it still presents two main drawbacks. First, it requires exemplar inputs, and its performance is highly sensitive to the selected examples. Second, like prior methods, it is prone to token explosion and substantial performance degradation when handling a large number of terms.
 
To address these limitations, we propose BoostTaxo, a boosting-style zero-shot method for taxonomy induction, which infers hierarchical relationships and their weights in a coarse-to-fine manner. The design of our method is primarily motivated by the following observations. First, LLMs are trained on massive textual corpora and thus encode rich semantic knowledge, which can be effectively elicited through appropriate prompt design\cite{brown2020language,wei2022chain}. Second, lightweight LLMs can substantially reduce computational cost, memory consumption, and inference latency, while still preserving considerable parametric knowledge and basic reasoning capability\cite{zhang2024tinyllama,srivastava2025towards}. These characteristics make them particularly well suited for large-scale candidates selection. Third, large-scale LLMs typically demonstrate stronger reasoning ability due to their increased parameter scale, enhanced representation capacity, and broader knowledge coverage, but this comes at the cost of substantially increased computational requirements\cite{sardana2023beyond,wei2022emergent}. 

As shown in Fig. \ref{fig1}, we propose a boosting-style zero-shot framework for taxonomy induction that identifies candidate parents for each term in a coarse-to-fine manner. First, a lightweight LLM (e.g., Qwen3-4B) is prompted to detect hypernymic relationships between term pairs, combined with definition matching, enabling the target term to gather information from all candidate terms. Next, based on the filtered candidate set, a large-scale LLM (e.g., GPT-4o) ranks and scores the selected candidates, allowing the target term to capture information from candidate terms at the same hierarchical level. Finally, structural features, including global structural features and local structural features  are incorporated into the prompt to guide the large-scale LLM in reasoning about and calibrating edge weights, thereby improving its awareness of the taxonomy structure. The main contributions of the paper are:
\begin{enumerate}

\item  We propose a boosting-style zero-shot framework for taxonomy induction, which identifies candidate parent terms for each target term in a coarse-to-fine manner without requiring task-specific training data. This design improves the generality and applicability of taxonomy induction in zero-shot settings.

\item We introduce a progressive collaboration between a lightweight LLM and a large-scale LLM, in which the former first filters candidate parents and the latter subsequently ranks and scores them, balancing efficiency and effectiveness.

\item We incorporate structure-aware reasoning based on structural features into large-scale LLMs to refine candidate parent weights, thereby producing more reliable candidate parents.

\item Extensive experiments on three public benchmark datasets demonstrate that our method achieves state-of-the-art performance.
\end{enumerate}

The remaining paper is structured as follows. Section \uppercase\expandafter{\romannumeral2} reviews related work. Section \uppercase\expandafter{\romannumeral3} presents the proposed methodology. Section \uppercase\expandafter{\romannumeral4} evaluates the model performance, conducts ablation studies, and provides further analysis. The final Section \uppercase\expandafter{\romannumeral5} concludes the paper and outlines directions for future research.

\section{RELATED WORK}
This review focuses on taxonomy induction and the application of large language models to knowledge extraction and reasoning.

\subsection {Taxonomy induction}
Taxonomy induction has evolved from pattern-based hypernym extraction to data-driven structure learning and, more recently, to pretrained and large language model based approaches. Early studies were largely founded on Hearst-style lexico-syntactic patterns, which identify explicit textual cues such as “X is a Y” or “Y such as X” to extract \textit{is-a} relations\cite{hearst1992automatic,kozareva2010semi,panchenko2016taxi,le2019inferring}. Building on this line, subsequent work introduced supervised and semi-supervised methods that combine heterogeneous evidence, web statistics, and structural constraints to infer full taxonomies rather than isolated hypernym pairs\cite{shwartz2016improving,snow2004learning}. Snow et al.\cite{snow2004learning} proposed a supervised framework that automatically learns dependency-path patterns from parsed text, replacing manually designed lexico-syntactic rules for hypernym discovery. They further combined hypernym evidence with coordinate-term information, showing that heterogeneous syntactic and structural cues can improve taxonomy induction beyond isolated pattern matches. Later, representation-based methods incorporated distributed semantics and global optimization, including word-embedding-based hierarchy learning, hypernym subsequence induction, reinforcement-learning-based end-to-end taxonomy construction, and unsupervised topic taxonomy generation, which improved robustness beyond manually designed patterns\cite{fu2014learning,tuan2016learning,mao2018end}. 

More recent studies have increasingly adopted pretrained language models, which use masked or prompted semantic knowledge to score parent–child relations and reconcile them into tree structures, enabling stronger generalization in low-resource or unseen domains. Chen et al.\cite{chen2021constructing} fine-tuned pretrained language models (e.g., BERT and RoBERTa) for pairwise parenthood prediction, using term-pair inputs of the form ``$v_i$ is a $v_j$'' to estimate whether one term is the direct parent of another. When web-retrieved glosses are available, the same parenthood prediction module is further fine-tuned on input sequences augmented with reordered glosses for both terms. Jain et al\cite{jain2022distilling} proposed two zero-shot taxonomy induction methods based on pretrained language models: MLM prompting, which predicts hypernyms through prompt completion, and LMScorer, which ranks candidate hypernyms by scoring the fluency of hypernymy-eliciting sentences. Both methods avoid task-specific fine-tuning and directly distill hypernymy knowledge from pretrained LMs. Most recently, LLMs have been explored for taxonomy induction in an in-context manner; for example, Chain-of-Layer\cite{zeng2024chain} iteratively prompts LLMs to expand taxonomies layer by layer and introduces an Ensemble-based Ranking Filter to mitigate hallucinated content. Its strong performance under limited examples highlights the great potential of LLMs for taxonomy construction.

\subsection {Large language models for knowledge extraction and reasoning}
LLMs have recently made significant advances in both knowledge extraction and reasoning, and have been widely applied to various downstream tasks. In knowledge extraction, recent studies have evolved from pipeline-based methods to generative frameworks that directly produce structured outputs, such as triples, events, and schema. REBEL reformulated relation extraction as an end-to-end triplet generation task \cite{cabot2021rebel}, while UIE further unified multiple information extraction tasks under a text-to-structure paradigm \cite{lu2022unified}. Subsequent studies extended this line of research to more settings, such as instruction-driven and on-demand extraction \cite{jiao2023instruct,qi2024adelie}, and also placed greater emphasis on more realistic evaluation protocols and robustness in generative extraction \cite{jiang2024genres,zhu2025towards}.

In reasoning, LLMs have shown strong capabilities in handling complex inference tasks. Their reasoning performance has been further improved by techniques such as chain-of-thought prompting, which elicits intermediate reasoning steps \cite{wei2022chain}, and has achieved remarkable success in mathematical reasoning tasks \cite{kojima2022large}. Tool-augmented methods further enhance reliability through interaction with external programs \cite{gao2023pal}. More recently, structured knowledge has been incorporated to support more grounded reasoning \cite{shen2026reason}.

\section{METHODOLOGY}
\subsection {Problem definition}
We define a taxonomy as a directed acyclic graph $\mathcal{T}=(\mathcal{V},\mathcal{E})$, where $\mathcal{V}$ is the set of terms and $\mathcal{E}$ is the set of directed edges encoding hierarchical relations among these terms. In taxonomy induction, the input consists of a predefined concept set $\mathcal{V}$, in which each term is expressed as either a single word or a short phrase. The task is to identify the hierarchical structure over these terms and construct the taxonomy $\mathcal{T}$ accordingly.

\begin{figure*}
  \begin{center}
  \includegraphics[scale=0.66]{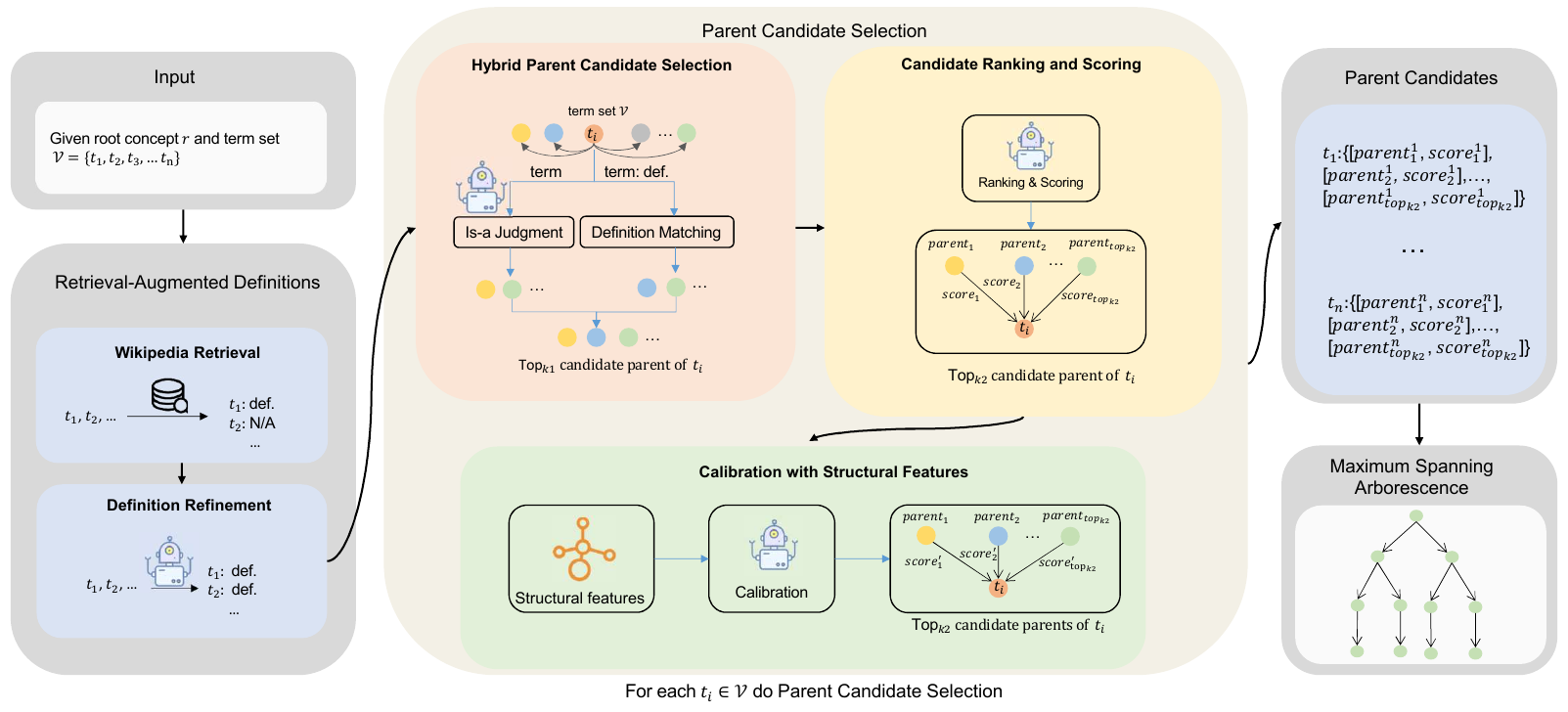}\\
 \caption{The boosting-style zero-shot framework for taxonomy induction. Starting from a root concept and a term set, the framework first enriches term semantics through retrieval-augmented definitions, which combine Wikipedia retrieval with definition refinement. Then, for each term, it performs parent candidate selection in three stages. First, Hybrid Parent Candidate Selection identifies a set of potential parents by jointly considering Is-A judgment and definition matching. Second, Candidate Ranking and Scoring ranks these candidates and assigns an initial score to each of them. Third, Calibration with Structural Features further adjusts these scores to improve their structural consistency and reliability. After obtaining parent candidates and calibrated scores for all terms, the final taxonomy is constructed by applying a maximum spanning arborescence algorithm.}\label{fig2}
  \end{center}
\end{figure*}

Taxonomy induction from scratch in the zero-shot setting remains challenging. To address this problem, we propose Boosting-Style Zero-Shot Taxonomy Induction with LLMs. As shown in Fig.~\ref{fig2}, the proposed framework consists of five core modules: Retrieval-Augmented Definitions, Hybrid Parent Candidate Selection, LLM-based Candidate Ranking and Scoring, LLM-based Score Calibration with Structural Features, and Maximum Spanning Arborescence.

\begin{enumerate}
    \item Retrieval-Augmented Definitions. It retrieves the definition of each term from Wikipedia and refines it using an LLM to enrich the semantic information of the terms.
    \item Hybrid Parent Candidate Selection. It combines \textit{is-a} judgment using a lightweight LLM with definition matching to retrieve the top$_{k1}$ candidate parents, enabling efficient, comprehensive, and low-cost identification of potential parent candidates.
    \item LLM-based Candidate Ranking and Scoring. It uses a large-scale LLM to rank and score the top$_{k2}$ candidates, ensuring that different parent candidate are compared on the same basis,  thereby enhancing the fairness and consistency of parent selection in a more fine-grained manner.
    \item LLM-based Score Calibration with Structural Features. It uses a large-scale LLM to further calibrate the candidate scores by incorporating local structural relation features and global structural features, thereby enhancing the reliability of parent selection.

    \item Maximum Spanning Arborescence. It employs the Chu–Liu/Edmonds algorithm\cite{edmonds1967optimum} to construct the final taxonomy based on the scores assigned to candidate parent-child relations.

\end{enumerate}

\subsection{Retrieval-Augmented Definitions}
\begin{figure}
  \begin{center}
  \includegraphics[scale=0.6]{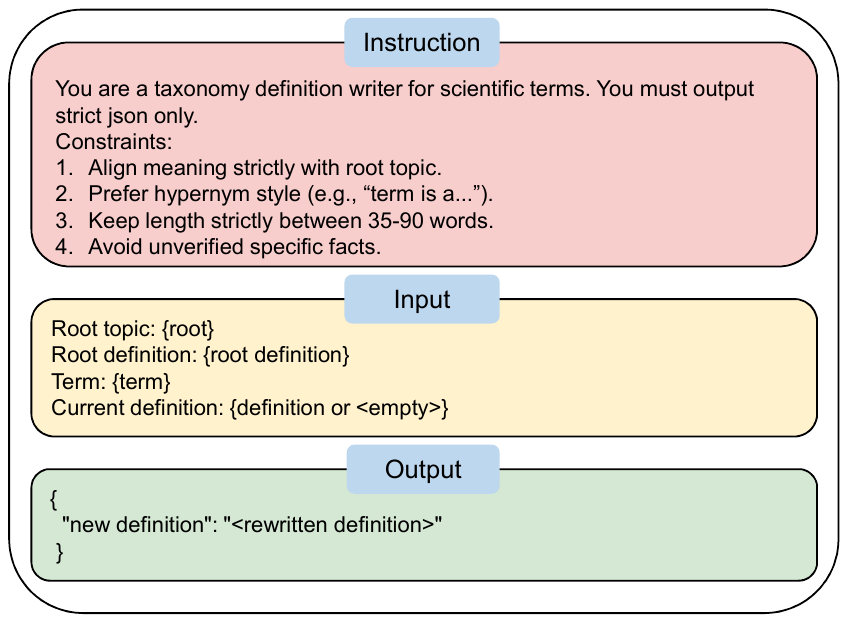}\\
 \caption{Prompt design for LLM-Based term definition refinement}\label{fig3}
  \end{center}
\end{figure}

The semantic information conveyed by an individual term is often limited. In cases of lexical ambiguity, such as Apple referring either to a fruit or to a smartphone brand, the intended meaning is highly dependent on context. To address this challenge, we propose a method for expanding the definition of a given term. Specifically, the method first retrieves candidate definitions through lexical matching from external knowledge bases such as Wikipedia. However, these retrieved definitions often exhibit several shortcomings, including missing entries, semantic ambiguity, excessively verbose wording, and overly brief descriptions. As shown in Fig. \ref{fig3}, to mitigate these issues, we design a structured prompt for LLM-based term definition refinement. The prompt takes the root topic, root definition, target term, and current definition as input, and constrains the model to generate a context-consistent, concise, and reliable rewritten definition in a unified JSON format.

\subsection{Hybrid Parent Candidate Selection}
Identifying parent candidates from a large set of candidate terms is a challenging task. On the one hand, supervised and semi-supervised learning methods can improve predictive accuracy, but they often exhibit limited generalization capability. On the other hand, traditional pattern-matching methods exhibit stronger generalization ability but often suffer from sparsity issues. To address these limitations, we propose a combined method that incorporates lightweight LLM-based \textit{is-a} relation judgment and term-definition matching. Our method is designed based on two key principles. First, lightweight LLMs are trained on large-scale corpora and therefore encode substantial semantic and world knowledge. In addition, their relatively low computational cost and fast inference speed make them particularly suitable for large-scale term processing scenarios. By designing appropriate prompt templates, the model can be guided to infer whether a specific hypernymy relation exists between two terms. Second, if two terms have a hypernymy relationship, their definitions often exhibit semantic inclusion or strong conceptual relevance, which can provide important evidence for parent–child identification.

Based on the first principle, and inspired by common Hearst-style lexico-syntactic patterns\cite{hearst1992automatic,zeng2024chain}, we adopt a set of templates $\mathcal{M}$ to capture potential hypernym--hyponym relations between a query term and an anchor term:
\begin{itemize}
    \item \texttt{<query> is a/an <anchor>}
    \item \texttt{<query> is a kind of <anchor>}
    \item \texttt{<query> is a type of <anchor>}
    \item \texttt{<query> is an example of <anchor>}
    \item \texttt{<anchor> such as <query>}
\end{itemize}

For each entity $q$ in the entity list, we pair it with each candidate anchor $a$ and feed the resulting query--anchor pair into a lightweight LLM(such as, Qwen3-4B) using the Hearst-style prompt templates defined above. In each template, $q$ is treated as the child term and $a$ as the candidate parent term, and the model is asked to determine whether an \textit{is-a} relation holds between them. Since each $(q,a)$ pair is evaluated once under each template, multiple judgments can be obtained from different lexico-syntactic perspectives.

Formally, for each template $m \in \mathcal{M}$, we define
$$
f_m(q,a)\in\{0,1\},
$$
where $f_m(q,a)=1$ indicates that the lightweight LLM judges $q$ and $a$ to satisfy an \textit{is-a} relation under template $m$, and $f_m(q,a)=0$ otherwise. The matching score between $q$ and $a$ is then defined as
$$
\mathrm{Score}(q,a \mid \mathcal{M})=\sum_{m \in \mathcal{M}} f_m(q,a).
$$

 We then aggregate these judgments across templates and use the resulting score to rank candidate terms for each $q$. The Top\_${k_\mathrm{isa}}$ mutually selected terms are finally retained as its parent candidates.

Based on the second principle, we further introduce a definition matching module to complement the LLM-based \textit{is-a} judgment. Specifically, for each entity in the same entity list, we construct a semantic representation by concatenating its term and definition. These representations are then encoded into dense embeddings using a sentence embedding model. After $\ell_2$ normalization, cosine similarity is computed between each query term and all candidate terms within the same list. For each term $q$, all remaining terms are ranked according to their semantic similarity to $q$, and the Top\_$k_{\mathrm{def}}$ most similar terms are retained as its definition-matching candidate set.

Formally, let $x_q$ denote the textual representation of term $q$, constructed by concatenating the term name and its definition, and let $E(\cdot)$ denote the sentence embedding model. The normalized embedding of $q$ is defined as
$$
\mathbf{e}_q = \frac{E(x_q)}{\|E(x_q)\|_2}.
$$
The semantic similarity between two terms $q$ and $a$ is then computed by cosine similarity:
$$
\mathrm{Sim}_{\mathrm{def}}(q,a)=\mathbf{e}_q^\top \mathbf{e}_a.
$$
Based on this score, all candidate terms are ranked for each query term $q$, and the Top\_$k_{\mathrm{def}}$ candidates are selected as the definition matching results.

Finally, for each term, we fuse the \textit{is-a} judgment candidate set with the definition matching candidate set through an order-preserving union operation. Specifically, the Top\_$k_{\mathrm{isa}}$ candidates produced by the LLM-based \textit{is-a} judgment are kept first, after which the previously Top\_$k_{\mathrm{def}}$ candidates from the definition matching results are appended in order. After duplicate removal and rank preservation, the Top$_{k1}$ candidates are retained as the final fused candidate set.

\subsection{Candidate Ranking and Scoring}
Although Hybrid Parent Candidate Selection can produce an initial set of candidate parents, it mainly relies on pairwise evidence between terms and does not explicitly model the relative differences among candidate parents at the same level. Consequently, the resulting candidate space is often overly broad, making it difficult to identify the true parent term. To address this problem, we further introduce a ranking and scoring stage to compare and refine the candidate parents. In this stage, we employ a large-scale LLM(such as, GPT-4o), whose stronger reasoning ability makes it more suitable than lightweight LLMs for fine-grained candidate discrimination.

\begin{figure}
  \begin{center}
  \includegraphics[scale=0.6]{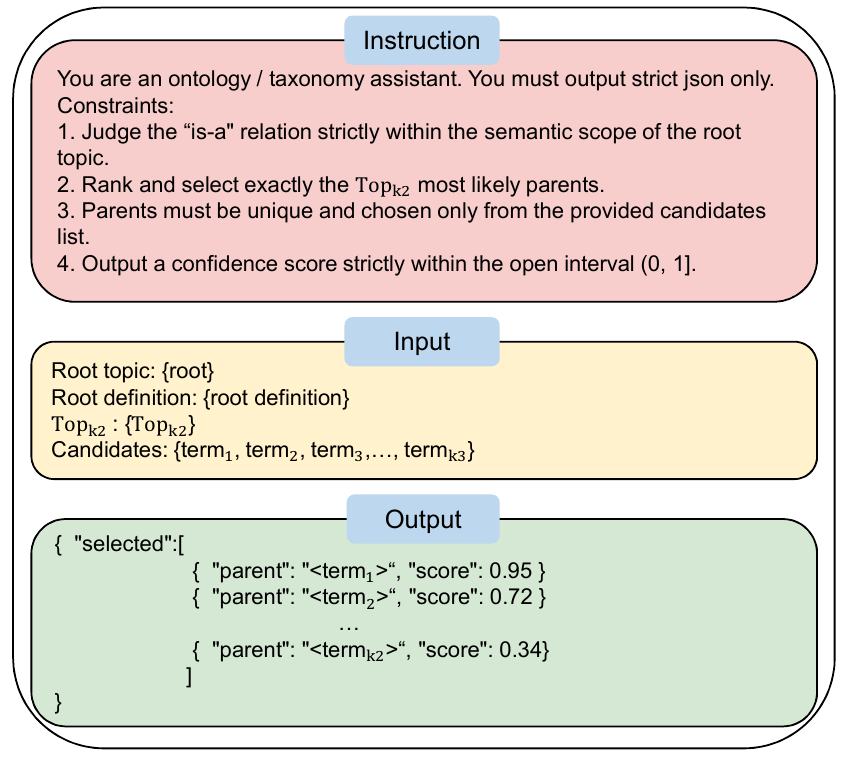}\\
 \caption{Prompt design for LLM-based candidate parent ranking and scoring}\label{fig4}
  \end{center}
\end{figure}

As shown in Fig. \ref{fig4}, the root topic, root definition, and the candidate parent set generated in the previous stage are provided to LLM. Within the semantic scope of the root topic, the model jointly ranks all candidates, selects exactly the Top$_{k2}$ most likely parents, and assigns each selected candidate a confidence score in $(0,1)$. By comparing candidates in a unified context rather than through independent pairwise judgments, the model can better capture subtle differences among semantically related candidates at the same level, thereby narrowing the candidate space and improving the identification of true parent terms.

\subsection{Score Calibration with Structural Features}
Although the ranking and scoring stage considers the relative differences among candidate parents at the same level, it still focuses primarily on semantic comparison within the candidate set and does not explicitly incorporate structural characteristics of the taxonomy, such as hierarchical depth and tree-level organization. As a result, the assigned scores may not fully reflect the true suitability of candidate parents in the overall taxonomy structure. To address this limitation, we further propose a LLM-based score calibration stage, which calibrates the candidate scores by incorporating additional structural features, thereby improving the reliability of parent candidate selection.

\subsubsection{Structural features construction}
Given the Top$_{k2}$ candidate parent set $P(t)$ for each child term $t$, we compute a set of structural and comparative features for each candidate edge $(t,p)$ to support the subsequent LLM-based penalty estimation. Let $s(t,p)\in[0,1]$ denote the normalized score of candidate edge $(t,p)$ obtained from the previous stage. These features can be grouped into two categories: global structural features and local structural relation features.

\noindent\textbf{(a) Global structure features}

\textbf{Popularity.} The popularity feature measures how frequently a candidate parent $p$ is selected by different child terms within the current tree. A higher popularity value suggests that $p$ may be a more general or overly broad concept. Let $\mathcal{C}$ denote the set of all unique child terms in the current tree, and let
\[
\mathrm{Child}(p)=\{\,t \mid (t,p)\text{ is a candidate edge}\,\}
\]
be the set of child terms that take $p$ as a candidate parent. Then the popularity feature is defined as
\[
\mathrm{pop}(p)=\frac{|\mathrm{Child}(p)|}{|\mathcal{C}|}.
\]

\textbf{Depth penalty.} The depth penalty feature measures how general a candidate parent is according to its position in the current candidate graph. We build a parent-to-child graph from all candidate edges in the current tree and perform a breadth-first search (BFS) from the root node $r$. For each node $p$, its depth $\mathrm{depth}(p)$ is defined as the shortest-path distance from $r$ to $p$, with $\mathrm{depth}(r)=0$. Shallower nodes are closer to the root and are therefore more likely to be overly general, while deeper nodes are usually more specific. Let
\[
D_{\max}=\max_{v}\mathrm{depth}(v)
\]
be the maximum reachable depth. Then the depth penalty is defined as
\[
\mathrm{depth\_pen}(p)=
\begin{cases}
\frac{1}{2}\left(1-\frac{\mathrm{depth}(p)}{D_{\max}}\right), & p \text{ is reachable from the root},\\[6pt]
0.25, & \text{otherwise}.
\end{cases}
\]

\noindent\textbf{(b) Local structural features}

\textbf{Margin.} The margin feature measures the relative competitive advantage of a candidate parent $p$ over the other candidate parents of the same child term $t$. A larger margin indicates that $p$ is more likely to be the best parent among the candidates. It is defined as
\[
\mathrm{margin}(t,p)=s(t,p)-\max_{q\in P(t),\,q\neq p}s(t,q).
\]

\textbf{Pullback.} The pullback feature measures whether a candidate parent $p$ is structurally supported by other candidate parents of the same child term $t$. If another candidate parent of $t$ also regards $p$ as its parent candidate, then $p$ may occupy a higher and more stable position in the local hierarchy. It is defined as
\[
\mathrm{pb}(t,p)=\frac{N_{t,p}}{|P(t)|-1},
\]
where $N_{t,p}$ is the number of candidate parents $f\in P(t)\setminus\{p\}$ such that $p\in P(f)$ and
\[
\min\big(s(f,p),\,s(t,f)\big)<s(t,p).
\]
For $|P(t)|\leq 1$, we set $\mathrm{pb}(t,p)=0$.

\textbf{Skip-level support.} The skip-level support feature measures whether a candidate parent $p$ is more likely to be an ancestor node rather than the direct parent of $t$. If there exists an intermediate node $m$ such that both $t\rightarrow m$ and $m\rightarrow p$ are strongly supported, then the edge $(t,p)$ may correspond to a skip-level relation. It is defined as
\[
\mathrm{skip\_support}(t,p)=
\max_{\substack{m\in P(t)\\ m\neq p,\; s(t,m)\ge s(t,p)-\delta}}
\min\big(s(t,m),\,s(m,p)\big),
\]
where $\delta$ is a tolerance margin, and the value is set to $0$ if no such intermediate node exists.

\textbf{Sibling cohesion.} The sibling cohesion feature measures the consistency of candidate children under the same parent candidate $p$. If the children associated with $p$ tend to have similar candidate parent sets, then $p$ is more likely to represent a coherent local grouping. Let
\[
S(p)=\{\,t \mid (t,p)\text{ is a candidate edge}\,\}
\]
denote the set of candidate children of $p$. Then the sibling cohesion is defined as
\[
\mathrm{coh}(t,p)=
\frac{1}{|S(p)\setminus\{t\}|}
\sum_{t'\in S(p)\setminus\{t\}}
J\big(P(t),P(t')\big),
\]
where
\[
J(A,B)=\frac{|A\cap B|}{|A\cup B|}
\]
is the Jaccard similarity. If $S(p)\setminus\{t\}=\varnothing$, then $\mathrm{coh}(t,p)=0$.

\subsubsection{LLM-Based Penalty Calculation with Structural Features}
Structural information characterizes the current tree from multiple perspectives. Fully leveraging this information to adjust edge scores can improve the reliability of candidate edges. However, manually designed weighting schemes for modeling the effects of different structural features on score calibration often struggle to adapt to diverse scenarios and generally exhibit limited generalization ability. Moreover, compared with directly estimating calibrated scores, modeling score penalties based on structural features is more interpretable and more naturally aligned with the calibration process. Motivated by the strong reasoning capabilities of large-scale LLMs, we therefore propose an LLM-based penalty calibration method that incorporates structural features.

\begin{figure}
  \begin{center}
  \includegraphics[scale=0.55]{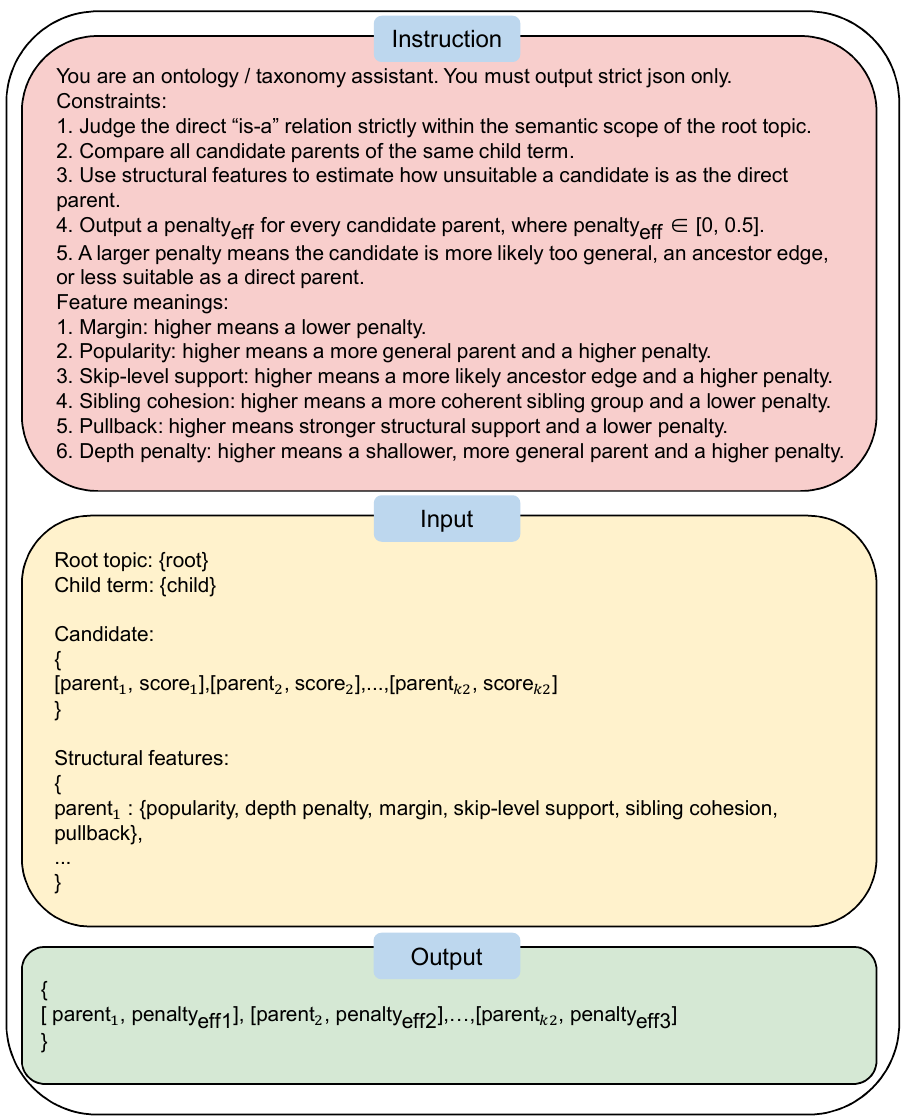}\\
 \caption{Prompt design for LLM-Based calibration of candidate parent scores}\label{fig5}
  \end{center}
\end{figure}

As shown in Fig. \ref{fig5}, given the child term, the LLM is prompted with the root topic, candidate parent terms, and a set of structural features associated with each  candidate parent. The instruction asks the LLM to evaluate is-a relation strictly within the semantic scope of the root topic, compare all candidate parents for the given child term, and estimate how suitable each candidate is as the direct parent. Specifically, the prompt incorporates six structural features, including margin, popularity, skip-level support, sibling cohesion, pullback, and depth penalty, which respectively capture confidence separation, parent generality, possible ancestor-edge tendency, sibling-group consistency, structural support from the existing tree, and shallowness of the candidate parent. Based on these features, the LLM outputs a penalty score for each candidate parent, where a larger penalty indicates that the candidate is more likely to be overly general, correspond to an ancestor edge, or otherwise be less suitable as the direct parent of the child term. The calibrated penalties are then used to revise the original edge scores, thereby improving parent selection and enhancing the overall structural reliability of the taxonomy.

\subsubsection{Overall score calibration}

Table \ref{tab:llm_rescore} summarizes the overall Procedure of the proposed LLM-based score calibration method. Starting from the candidate edge set generated in the previous stage, the method first removes unreliable mutual edges and normalizes the remaining edge scores. It then constructs candidate parent mappings and the induced parent-to-children graph, from which structural features are computed for each candidate edge, including margin, parent popularity, pullback support, skip-level support, sibling cohesion, and depth-related information derived from the root. For each child term, all candidate parents, together with these structural features, are submitted to the LLM, which predicts an effective penalty score for each candidate parent. The normalized edge scores are then re-calibrated using these penalties to obtain a refined candidate edge set. In this way, the proposed method combines comparative semantic evidence with structural features in a unified calibration framework, thereby improving the reliability of parent selection.

\begin{table}[t]
\centering
\caption{LLM-based score calibration algorithm}
\label{tab:llm_rescore}
\begin{tabular}{p{0.95\linewidth}}
\toprule
\textbf{Input:} Candidate edges $E$, root node $r$, threshold $\tau_m$, LLM \\
\textbf{Output:} Adjusted candidate edges $E'$ \\
\midrule

\hangindent=1.2em \hangafter=1
1. Filter mutual edges in $E$ with threshold $\tau_m$ to obtain the filtered edge set $E_f$. \\

\hangindent=1.2em \hangafter=1
2. Normalize the scores of edges in $E_f$ to obtain $s(t,p)\in[0,1]$. \\

\hangindent=1.2em \hangafter=1
3. Build candidate mappings $P(t)$ and the parent-to-children graph from $E_f$. \\

\hangindent=1.2em \hangafter=1
4. For each edge $(t,p)\in E_f$, compute $\mathrm{margin}(t,p)$, $\mathrm{pop}(p)$, $\mathrm{pb}(t,p)$, $\mathrm{skip\_support}(t,p)$, $\mathrm{coh}(t,p)$, and $\mathrm{depth\_pen}(p)$. \\

\hangindent=1.2em \hangafter=1
5. For each child term $t$, query the LLM with all candidate parents in $P(t)$ and obtain $\mathrm{penalty}_{\mathrm{eff}}(t,p)\in[0,0.5]$ for each $p\in P(t)$. \\

\hangindent=1.2em \hangafter=1
6. For each edge $(t,p)\in E_f$, update its score by $s'(t,p)=s(t,p)*(1-\mathrm{penalty}_{\mathrm{eff}}(t,p))$. \\

\hangindent=1.2em \hangafter=1
7. Return the adjusted edge set $E'=\{(t,p,s'(t,p)) \mid (t,p)\in E_f\}$. \\

\bottomrule
\end{tabular}
\end{table}

\subsection{Maximum Spanning Arborescence}
Given the child–parent pairs and their corresponding weights, we formulate taxonomy construction as a maximum spanning arborescence problem under taxonomy constraints. A spanning arborescence is similar to a spanning tree, but it applies to directed graphs. Therefore, We  use the Edmonds algorithm\cite{edmonds1967optimum} to obtain the final taxonomy, ensuring that all constraints are satisfied and the taxonomy keeps a tree structure.

\section{EXPERIMENTS AND EVALUATIONS}
\subsection {Data set}
The WordNet dataset comprises 761 non-overlapping sub-taxonomies extracted from WordNet, each with 11 to 50 entities. Following the standard split, the dataset is divided into training, development, and test sets, containing 533, 114, and 114 taxonomies, respectively. DBLP is constructed from 156,000 computer science paper abstracts, and its sub-taxonomies range from 80 to 120 entities. SemEval2016-Sci is derived from the taxonomy induction shared task in SemEval 2016, and its sub-taxonomies also range from 80 to 120 entities. For all datasets, we adopt the same test split as in the previous study\cite{zeng2024chain}.

\subsection {Benchmark models and evaluation metrics}
We evaluate our approach against several state-of-the-art methods on three widely used benchmark datasets: WordNet, DBLP, and SemEval2016-Sci. The comparison includes the following representative methods:

\begin{itemize}
    \item \textbf{Graph2Taxo}\cite{shang2020taxonomy}: This method leverages cross-domain graph structures and employs constraint-based directed acyclic graph (DAG) learning for taxonomy induction.
    
    \item \textbf{CTP}\cite{chen2021constructing}: This method fine-tunes a RoBERTa model or direct use the Llama-2-7B to predict the likelihood of parent-child pairs and then integrates these predictions into a graph using a maximum spanning tree algorithm for accurate taxonomy induction.
    
    \item \textbf{RestrictMLM}\cite{jain2022distilling}: This method uses a cloze-style, or fill-in-the-blank, formulation to extract \textit{is-a} relational knowledge from BERT. 
    
    \item \textbf{LMScore}\cite{jain2022distilling}: This method formulates taxonomy induction as a sentence scoring task based on GPT-2, evaluating the fluency of sentences that express parent-child relations.
    
    \item \textbf{TaxonomyGPT}\cite{chen2023prompting}: This method frames taxonomy induction as a conditional text generation task. It represents the generated taxonomy as a set of sentences, each describing a parent-child relation in the output taxonomy.

     \item \textbf{Chain-of-Layer}\cite{zeng2024chain}: This method takes all candidate terms as input simultaneously and decomposes taxonomy induction into a layer-wise process, in which relevant candidate entities are selected at each level and the taxonomy is progressively constructed from top to bottom.

\end{itemize}

To ensure a fair and comprehensive evaluation, we measure model performance using two complementary metrics: \textbf{Ancestor-F1} and \textbf{Edge-F1}. The definitions of these metrics are given below.

\textbf{Ancestor-F1.} This metric evaluates the consistency of ancestor--descendant relations between the predicted taxonomy and the ground-truth taxonomy. Formally, the ancestor precision, recall, and F1-score are defined as:
\begin{equation}
P_a = \frac{\left| \mathrm{is\text{-}ancestor}_{\mathrm{pred}} \cap \mathrm{is\text{-}ancestor}_{\mathrm{gold}} \right|}{\left| \mathrm{is\text{-}ancestor}_{\mathrm{pred}} \right|}
\end{equation}

\begin{equation}
R_a = \frac{\left| \mathrm{is\text{-}ancestor}_{\mathrm{pred}} \cap \mathrm{is\text{-}ancestor}_{\mathrm{gold}} \right|}{\left| \mathrm{is\text{-}ancestor}_{\mathrm{gold}} \right|}
\end{equation}

\begin{equation}
F1_a = \frac{2 P_a R_a}{P_a + R_a}
\end{equation}
where $P_a$, $R_a$, and $F1_a$ represent the ancestor precision, recall, and F1-score, respectively.

\textbf{Edge-F1.} This metric is stricter than Ancestor-F1 because it directly compares the predicted edges with the gold-standard edges. The corresponding edge precision, recall, and F1-score are denoted by $P_e$, $R_e$, and $F1_e$, respectively.

\subsection {Experimental setup}
In the retrieval-augmented definition stage, ChatGPT-4o is employed for definition refinement. In the hybrid parent candidate selection stage, Qwen3-4B is used as the lightweight LLM for \textit{is-a} detection, while all-mpnet-base-v2 is adopted for definition matching; Top$_{k_{\mathrm{isa}}}$ and Top$_{k_{\mathrm{def}}}$ are set to 10 and 5, respectively. In the LLM-based candidate ranking and scoring stage, ChatGPT-4o is used as the large-scale LLM for candidate ranking and scoring, with Top$_{k2}$ set to 3. In the LLM-based score calibration with structural features stage, ChatGPT-4o is further used as the large-scale LLM for candidate score calibration.

\subsection {Results}
Table \uppercase\expandafter{\romannumeral2} presents the performance comparison on the WordNet, DBLP, and SemEval-Sci datasets. Overall, the proposed BoostTaxo method achieves competitive or superior results compared with both supervised fine-tuning methods and prompting-based LLM baselines under zero-shot and few-shot settings. In particular, under the zero-shot setting, BoostTaxo (GPT-4o) achieves the best $F1_a$ on WordNet (82.79) and SemEval-Sci (78.39), as well as the best $F1_e$ on all three datasets, namely 63.50 on WordNet, 48.55 on DBLP, and 60.88 on SemEval-Sci. These results indicate that the proposed method can effectively identify more reliable parent--child relations while preserving stronger global hierarchical consistency. The improvement can be attributed to the coarse-to-fine design of the framework, where lightweight LLM-based candidate filtering reduces the search space, large-scale LLM-based ranking improves fine-grained parent discrimination, and structure-aware score calibration further enhances the reliability of edge selection by incorporating global and local structural features.

Compared with supervised fine-tuning methods such as Graph2Taxo and CTP, the proposed method demonstrates stronger generalization ability in zero-shot taxonomy induction. Although Graph2Taxo achieves the highest $P_e$ on WordNet and SemEval-Sci, its relatively low recall leads to lower overall F1 scores, suggesting that it tends to make conservative predictions. A similar pattern can also be observed in some other baselines, where high precision is achieved at the cost of insufficient recall, thus limiting the completeness of the induced taxonomy. In contrast, BoostTaxo (GPT-4o) maintains a better balance between precision and recall, especially on WordNet and SemEval-Sci, which results in stronger overall performance in both ancestor-level and edge-level evaluation.

Compared with prompting-based LLM methods, the advantages of the proposed method are also clear. TaxonomyGPT and CoL achieve promising results in the few-shot setting, especially when stronger LLMs are used, showing that in-context demonstrations can partially improve taxonomy construction. However, these methods still rely heavily on exemplar quality and often struggle to balance efficiency, structural consistency, and scalability. In the zero-shot setting, CoL-zero (GPT-4) already provides strong results, but BoostTaxo (GPT-4o) further surpasses it on the most important overall metrics, including $F1_a$ on WordNet and SemEval-Sci and $F1_e$ on all three datasets. This suggests that explicitly decomposing taxonomy induction into candidate generation, ranking, and calibration is more effective than directly relying on holistic prompting alone. In addition, CoL-zero (GPT-4o) attains extremely high precision on DBLP, but its very low recall causes a substantial drop in F1, indicating that overly selective predictions may hurt taxonomy completeness. By contrast, the proposed method achieves much stronger recall and a more balanced overall performance, demonstrating the effectiveness of the boosting-style reasoning and calibration strategy.

\begin{table*}[t]
\centering
\caption{Performance comparison on WordNet, DBLP, and SemEval-Sci datasets.}
\label{tab:main_results}
\scriptsize
\setlength{\tabcolsep}{4pt}
\renewcommand{\arraystretch}{1.12}

\begin{tabular}{lcccccc cccccc cccccc}
\toprule
\multirow{2}{*}{Model} 
& \multicolumn{6}{c}{WordNet} 
& \multicolumn{6}{c}{DBLP} 
& \multicolumn{6}{c}{SemEval-Sci} \\
\cmidrule(lr){2-7} \cmidrule(lr){8-13} \cmidrule(lr){14-19}
& $P_a$ & $R_a$ & $F1_a$ & $P_e$ & $R_e$ & $F1_e$
& $P_a$ & $R_a$ & $F1_a$ & $P_e$ & $R_e$ & $F1_e$
& $P_a$ & $R_a$ & $F1_a$ & $P_e$ & $R_e$ & $F1_e$ \\
\midrule

\multicolumn{19}{c}{\textit{Supervised Fine-tuning}} \\
\midrule
Graph2Taxo
& 79.20 & 47.80 & 59.60 & \textbf{75.60} & 37.00 & 49.70
& 47.85 & 30.23 & 37.05 & 46.63 & 28.49 & 35.37
& 82.45 & 36.15 & 50.27 & \textbf{79.37} & 34.52 & 46.87 \\
CTP
& 69.30 & 66.20 & 66.70 & 53.30 & 49.80 & 51.50
& 45.62 & 41.39 & 43.40 & 38.21 & 33.73 & 35.83
& 52.41 & 33.88 & 41.16 & 31.18 & 29.42 & 30.27 \\
CTP-Llama-2-7B
& 73.48 & 70.02 & 71.71 & 55.42 & 51.98 & 53.64
& 48.73 & 39.88 & 43.86 & 44.39 & 35.81 & 39.64
& 61.98 & {54.09} & 57.77 & 48.33 & 41.92 & 44.90 \\
\midrule

\multicolumn{19}{c}{\textit{5-shot Setting}} \\
\midrule
TaxonomyGPT (GPT-4) 
& 53.09 & 31.84 & 39.07 & 39.59 & 36.84 & 38.01
& 28.98 & 14.40 & 17.15 & 34.27 & 22.17 & 25.97
& 53.09 & 31.84 & 39.07 & 39.59 & 36.84 & 38.01 \\
CoL (GPT-4)
& \textbf{90.60} & 73.07 & 79.62 & 59.57 & 57.10 & 57.73
& 79.95 & 63.06 & {68.82} & 55.07 & 44.27 & 47.96
& 91.23 & 48.16 & 62.69 & 59.60 & 46.03 & 51.59 \\
CoL (GPT-4o)
& {89.95} & {76.50} & {82.14} & 62.52 & {61.47} & {61.97}
& 79.65 & {71.16} & \textbf{75.16} & {58.42} & \textbf{52.68} & \textbf{55.38}
& \textbf{95.58} & 48.90 & {64.60} & 60.04 & {49.64} & {54.29} \\
\midrule

\multicolumn{19}{c}{\textit{Zero-shot Setting}} \\
\midrule
RestrictMLM 
& 23.23 & 25.69 & 24.09 & 24.17 & 25.65 & 24.89
& - & - & - & - & - & -
& 63.33 & 47.85 & 54.44 & 45.79 & 46.19 & 45.99 \\
LMScorer
& 37.50 & 47.64 & 41.59 & 36.27 & 38.48 & 37.34
& 17.14 & 21.54 & 19.04 & 25.84 & 26.12 & 25.98
& 48.80 & 33.24 & 39.51 & 42.20 & 42.58 & 42.39 \\
CoL-zero (GPT-4)
& 89.71 & 71.39 & 78.31 & 58.93 & 55.18 & 56.41
& {80.88} & 54.25 & 57.21 & 40.15 & 35.88 & 37.81
& {94.99} & 45.83 & 61.66 & {62.33} & 45.55 & 52.44 \\
CoL-zero (GPT-4o)
& 75.03 & 61.28 & 65.97 & 53.06 & 48.83 & 50.08
& \textbf{94.69} & 12.01 & 21.31 & \textbf{85.71} & 12.03 & 21.09
& 57.67 & 21.88 & 27.96 & 34.76 & 19.05 & 21.40 \\
BoostTaxo (GPT-4o)
& 84.29 & \textbf{81.34} & \textbf{82.79} & {63.57} & \textbf{63.43} & \textbf{63.50}
& 59.02 & \textbf{80.81} & 68.21 & 48.55 & {48.55} & {48.55}
& 88.11 & \textbf{70.60} & \textbf{78.39} & 60.88 & \textbf{60.88} & \textbf{60.88} \\
\bottomrule
\end{tabular}
\end{table*}

\subsection {Ablation study}
In this section, we conduct an ablation study to evaluate the effects of Hybrid Parent Candidate Selection, LLM-based Score Calibration with Structural Features, and Top$_{k2}$ selection.

\subsubsection{Impact of Hybrid Parent Candidate Selection}
To evaluate the impact of Hybrid Parent Candidate Selection (HPCS), we compare the performance of the proposed framework with and without HPCS on the WordNet, DBLP, and SemEval-Sci datasets. As shown in Table \ref{tab:ablation_hpcg}, HPCS consistently improves ancestor-level performance across all three datasets, with $F1_a$ increasing from 76.47 to 79.81 on WordNet, from 48.80 to 54.07 on DBLP, and from 72.39 to 77.34 on SemEval-Sci. It also improves edge-level $F1_e$ on WordNet and SemEval-Sci, from 57.33 to 61.20 and from 54.03 to 57.88, respectively.

Without HPCS, all other terms in the taxonomy are directly treated as candidate parents for a given entity. As the number of entities grows, this strategy substantially enlarges the candidate space and introduces more noisy or weakly related parents. This effect is especially pronounced on larger taxonomies such as DBLP and SemEval-Sci, whose sub-taxonomies contain 80 to 120 entities. As a result, although the framework without HPCS tends to achieve high recall, it often suffers from relatively low precision, as shown by the DBLP results. In contrast, HPCS first applies lightweight LLM-based \textit{is-a} checking and definition matching to filter out implausible parent candidates before ranking, making the candidate generation process more focused and informative. This helps reduce interference from irrelevant terms and leads to a better balance between precision and recall, thereby improving the overall structural quality of the induced taxonomy.

\begin{table*}[t]
\centering
\caption{Ablation study on the effect of Hybrid Parent Candidate Selection across WordNet, DBLP, and SemEval-Sci datasets.}
\label{tab:ablation_hpcg}
\scriptsize
\setlength{\tabcolsep}{4pt}
\renewcommand{\arraystretch}{1.12}

\begin{threeparttable}
\begin{tabular}{lcccccc cccccc cccccc}
\toprule
\multirow{2}{*}{Setting}
& \multicolumn{6}{c}{WordNet}
& \multicolumn{6}{c}{DBLP}
& \multicolumn{6}{c}{SemEval-Sci} \\
\cmidrule(lr){2-7} \cmidrule(lr){8-13} \cmidrule(lr){14-19}
& $P_a$ & $R_a$ & $F1_a$ & $P_e$ & $R_e$ & $F1_e$
& $P_a$ & $R_a$ & $F1_a$ & $P_e$ & $R_e$ & $F1_e$
& $P_a$ & $R_a$ & $F1_a$ & $P_e$ & $R_e$ & $F1_e$ \\
\midrule
w/o HPCS
& 71.60 & 82.06 & 76.47 & 57.38 & 57.28 & 57.33
& 34.27 & 84.74 & 48.80 & 45.66 & 45.66 & 45.66
& 90.28 & 60.41 & 72.39 & 54.08 & 53.98 & 54.03 \\
w/ HPCS
& 79.00 & 80.65 & 79.81 & 61.21 & 61.19 & 61.20
& 39.55 & 85.41 & 54.07 & 43.21 & 43.21 & 43.21
& 93.88 & 65.76 & 77.34 & 57.88 & 57.88 & 57.88 \\
\bottomrule
\end{tabular}

\begin{tablenotes}[flushleft]
\footnotesize
\item HPCS denotes \textit{Hybrid Parent Candidate Selection}.
\end{tablenotes}
\end{threeparttable}
\end{table*}

\subsubsection{Impact of the LLM-based Score Calibration with Structural Features}
To evaluate the effect of LLM-based Score Calibration with Structural Features (LSC-SF), we compare the proposed framework with and without this module on the WordNet, DBLP, and SemEval-Sci datasets. As shown in Table \ref{tab:ablation_lscsf}, incorporating LSC-SF consistently improves the overall performance across all three datasets. Specifically, $F1_a$ increases from 79.81 to 82.79 on WordNet, from 54.07 to 68.21 on DBLP, and from 77.34 to 78.39 on SemEval-Sci. Edge-level $F1_e$ also improves from 61.20 to 63.50 on WordNet, from 43.21 to 48.55 on DBLP, and from 57.88 to 60.88 on SemEval-Sci.These results show that LSC-SF effectively improves the reliability of parent selection by leveraging both global and local structural features. Without this module, the framework already achieves relatively strong recall, but its precision remains limited, especially on DBLP. After introducing LSC-SF, the balance between precision and recall becomes notably better, leading to more accurate and structurally consistent taxonomies. This demonstrates that structure-aware score calibration plays an important role in refining candidate rankings and enhancing the overall quality of taxonomy induction.

\begin{table*}[t]
\centering
\caption{Ablation study on the effect of LLM-based Score Calibration with Structural Features across WordNet, DBLP, and SemEval-Sci datasets.}
\label{tab:ablation_lscsf}
\scriptsize
\setlength{\tabcolsep}{4pt}
\renewcommand{\arraystretch}{1.12}

\begin{threeparttable}
\begin{tabular}{lcccccc cccccc cccccc}
\toprule
\multirow{2}{*}{Setting}
& \multicolumn{6}{c}{WordNet}
& \multicolumn{6}{c}{DBLP}
& \multicolumn{6}{c}{SemEval-Sci} \\
\cmidrule(lr){2-7} \cmidrule(lr){8-13} \cmidrule(lr){14-19}
& $P_a$ & $R_a$ & $F1_a$ & $P_e$ & $R_e$ & $F1_e$
& $P_a$ & $R_a$ & $F1_a$ & $P_e$ & $R_e$ & $F1_e$
& $P_a$ & $R_a$ & $F1_a$ & $P_e$ & $R_e$ & $F1_e$ \\
\midrule
w/o LSC-SF
& 79.00 & 80.65 & 79.81 & 61.21 & 61.19 & 61.20
& 39.55 & 85.41 & 54.07 & 43.21 & 43.21 & 43.21
& 93.88 & 65.76 & 77.34 & 57.88 & 57.88 & 57.88 \\
w/ LSC-SF
& 84.29 & 81.34 & 82.79 & 63.57 & 63.43 & 63.50
& 59.02 & 80.81 & 68.21 & 48.55 & 48.55 & 48.55
& 88.11 & 70.60 & 78.39 & 60.88 & 60.88 & 60.88 \\
\bottomrule
\end{tabular}

\begin{tablenotes}[flushleft]
\footnotesize
\item LSC-SF denotes \textit{LLM-based Score Calibration with Structural Features}.
\end{tablenotes}
\end{threeparttable}
\end{table*}

\subsubsection{Impact of the Top$_{k2}$ selection}
To evaluate the impact of the Top$_{k2}$ selection, we compare different values of $top_{k2}$ on the WordNet, DBLP, and SemEval-Sci datasets. As shown in Table \ref{tab:ablation_topk2}, different choices of $top_{k2}$ lead to different results. When $top_{k2}=1$, the framework achieves the best performance on WordNet, with $F1_a$ and $F1_e$ reaching 82.65 and 63.69, respectively. This is likely because WordNet contains relatively small sub-taxonomies, each with only 11 to 50 entities, so a smaller candidate set is often sufficient to identify the correct parent while reducing noise. However, its performance on SemEval-Sci is lower than that of $top_{k2}=3$, and the edge-level performance on DBLP is also weaker.

When $top_{k2}=5$, recall further increases on DBLP and SemEval-Sci, but precision drops noticeably, leading to lower overall $F1_a$. This trend is more evident on larger taxonomies, where each sub-taxonomy contains 80 to 120 entities, as a larger $top_{k2}$ introduces more candidate parents but also more irrelevant or misleading ones. In contrast, $top_{k2}=3$ provides the best overall balance across datasets, achieving the highest $F1_a$ on SemEval-Sci and more stable performance on DBLP. These results indicate that the optimal choice of $top_{k2}$ is related to the scale of the taxonomy: a smaller value is more suitable for compact taxonomies such as WordNet, while a moderate value is more effective for larger and more complex taxonomies such as DBLP and SemEval-Sci. Therefore, setting $top_{k2}=3$ provides a more suitable balance between candidate diversity and prediction reliability.
\begin{table*}[t]
\centering
\caption{Ablation study on the choice of $top_{k2}$ across WordNet, DBLP, and SemEval-Sci datasets.}
\label{tab:ablation_topk2}
\scriptsize
\setlength{\tabcolsep}{4pt}
\renewcommand{\arraystretch}{1.12}

\begin{threeparttable}
\begin{tabular}{lcccccc cccccc cccccc}
\toprule
\multirow{2}{*}{Setting}
& \multicolumn{6}{c}{WordNet}
& \multicolumn{6}{c}{DBLP}
& \multicolumn{6}{c}{SemEval-Sci} \\
\cmidrule(lr){2-7} \cmidrule(lr){8-13} \cmidrule(lr){14-19}
& $P_a$ & $R_a$ & $F1_a$ & $P_e$ & $R_e$ & $F1_e$
& $P_a$ & $R_a$ & $F1_a$ & $P_e$ & $R_e$ & $F1_e$
& $P_a$ & $R_a$ & $F1_a$ & $P_e$ & $R_e$ & $F1_e$ \\
\midrule
$top_{k2}=1$
& 82.55 & 82.75 & 82.65 & 63.71 & 63.66 & 63.69
& 59.86 & 76.66 & 67.22 & 40.53 & 40.53 & 40.53
& 90.59 & 63.59 & 74.73 & 56.64 & 56.64 & 56.64 \\
$top_{k2}=3$
& 79.00 & 80.65 & 79.81 & 61.21 & 61.19 & 61.20
& 39.55 & 85.41 & 54.07 & 43.21 & 43.21 & 43.21
& 93.88 & 65.76 & 77.34 & 57.88 & 57.88 & 57.88 \\
$top_{k2}=5$
& 79.17 & 79.98 & 79.57 & 60.72 & 60.69 & 60.71
& 33.28 & 85.86 & 47.96 & 43.43 & 43.43 & 43.43
& 94.41 & 63.89 & 76.21 & 57.35 & 57.35 & 57.35 \\
\bottomrule
\end{tabular}
\end{threeparttable}
\end{table*}

\subsection {Impact of Different Large-Scale LLMs}
Table~\ref{tab:ablation_llm} shows that the proposed boosting-style zero-shot taxonomy induction framework is effective across different backbone LLMs, while the performance still varies with model capability. Overall, GPT-4o achieves the most stable and strongest results on DBLP,SemEval-Sci and WordNet, indicating better semantic understanding and hierarchical decision-making in complex domains. Qwen3-14B and Qwen3-30B also demonstrate competitive performance, especially on WordNet, although their improvements are less consistent across datasets. GPT-4mini yields reasonable ancestor-level performance, but remains weaker on edge-level prediction. These results suggest that stronger LLMs can provide more reliable support for both parent selection and taxonomy structure calibration in large-scale zero-shot taxonomy induction.

\begin{table*}[t]
\centering
\caption{Ablation study results across WordNet, DBLP, and SemEval-Sci datasets using Micro average scores.}
\label{tab:ablation_llm}
\scriptsize
\setlength{\tabcolsep}{4pt}
\renewcommand{\arraystretch}{1.12}

\begin{threeparttable}
\begin{tabular}{lcccccc cccccc cccccc}
\toprule
\multirow{2}{*}{Model}
& \multicolumn{6}{c}{WordNet}
& \multicolumn{6}{c}{DBLP}
& \multicolumn{6}{c}{SemEval-Sci} \\
\cmidrule(lr){2-7} \cmidrule(lr){8-13} \cmidrule(lr){14-19}
& $P_a$ & $R_a$ & $F1_a$ & $P_e$ & $R_e$ & $F1_e$
& $P_a$ & $R_a$ & $F1_a$ & $P_e$ & $R_e$ & $F1_e$
& $P_a$ & $R_a$ & $F1_a$ & $P_e$ & $R_e$ & $F1_e$ \\
\midrule
Qwen3-14B
& 85.30 & 75.04 & 79.84 & 55.46 & 55.21 & 55.34
& 51.83 & 76.21 & 61.70 & 39.73 & 39.64 & 39.69
& 85.38 & 65.36 & 74.04 & 53.37 & 53.27 & 53.32 \\
Qwen3-30B
& 77.10 & 80.27 & 78.65 & 55.77 & 55.62 & 55.69
& 53.72 & 79.46 & 64.10 & 46.41 & 46.10 & 46.26
& 76.57 & 70.85 & 73.60 & 57.35 & 57.35 & 57.35 \\
GPT-4mini
& 78.82 & 70.27 & 74.30 & 46.56 & 46.27 & 46.42
& 53.78 & 78.23 & 63.74 & 42.32 & 42.32 & 42.32
& 91.56 & 63.44 & 74.95 & 47.43 & 47.43 & 47.43 \\
GPT-4o
& 84.29 & 81.34 & 82.79 & 63.57 & 63.43 & 63.50
& 59.02 & 80.81 & 68.21 & 48.55 & 48.55 & 48.55
& 88.11 & 70.60 & 78.39 & 60.88 & 60.88 & 60.88 \\
\bottomrule
\end{tabular}

\end{threeparttable}
\end{table*}

\subsection {Example Results and Failure Cases}
\begin{figure*}
  \begin{center}
  \includegraphics[scale=0.6]{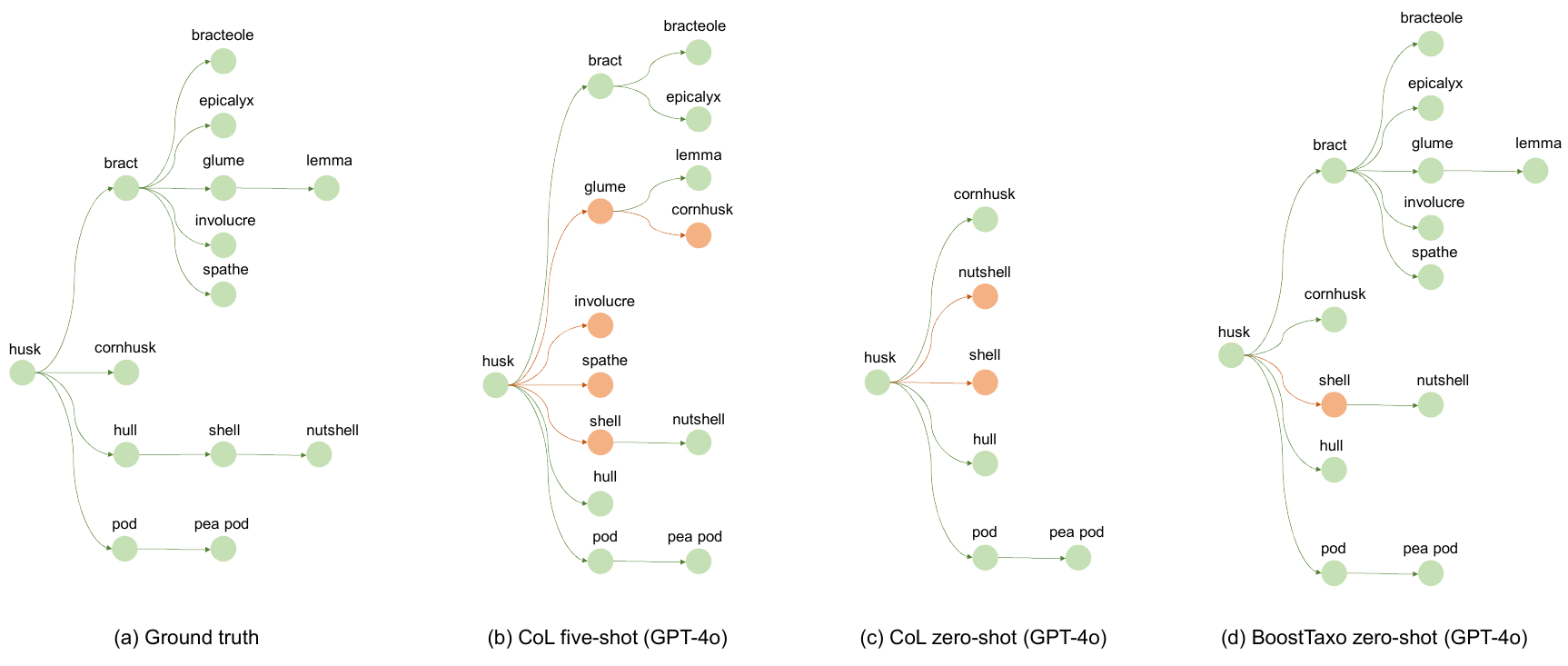}\\
 \caption{Comparison of taxonomy construction results across different methods.(a) is the ground truth taxonomy, and (b)–(d) show the taxonomies generated by GPT-4o using CoL five-shot, CoL zero-shot, and BoostTaxo zero-shot, respectively. Green indicates correct nodes and edges, while orange indicates incorrect ones. Among the three settings, BoostTaxo zero-shot yields the structure that is closest to the ground truth.}\label{fig6}
  \end{center}
\end{figure*}

Fig.~\ref{fig6} shows a example resul comparing different methods for taxonomy induction. Compared with the ground-truth taxonomy in Fig.~\ref{fig6}(a), CoL zero-shot in Fig.~\ref{fig6}(c) misses an important branch and produces incorrect local relations, while CoL five-shot in Fig.~\ref{fig6}(b) recovers more terms but still suffers from several attachment errors. In contrast, our BoostTaxo zero-shot method in Fig.~\ref{fig6}(d) generates a structure that is most consistent with the ground truth, indicating a stronger ability to capture global hierarchical relations.

\begin{figure*}
  \begin{center}
  \includegraphics[scale=0.65]{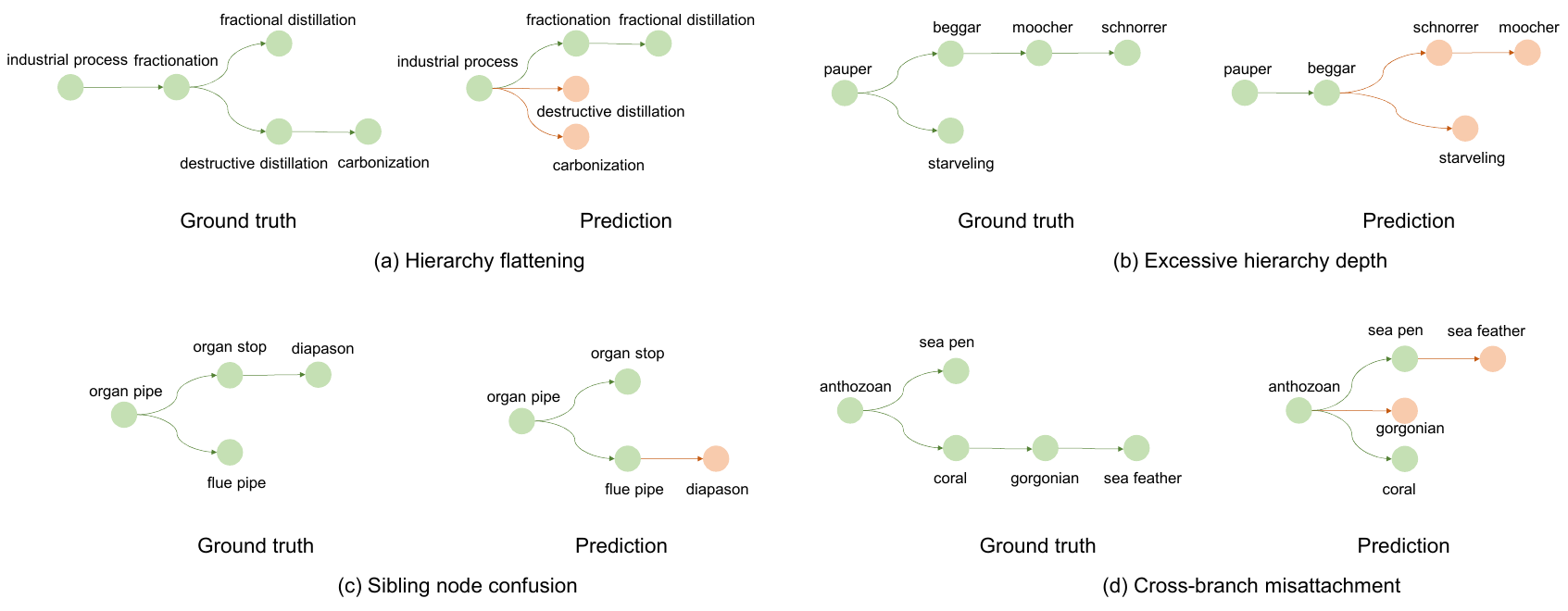}\\
 \caption{Representative failure cases in taxonomy construction. In each subfigure, the left side shows the ground-truth taxonomy and the right side shows the predicted taxonomy. Green indicates correct terms and edges, while orange indicates incorrect ones. (a) shows hierarchy flattening, where intermediate levels are removed and descendants are directly attached to higher-level terms. (b) shows excessive hierarchy depth, where extra intermediate terms are incorrectly introduced. (c) shows sibling node confusion, where a term is attached to an incorrect sibling branch. (d) shows cross-branch misattachment, where a term is assigned to the wrong semantic branch.}
 \label{fig7}
  \end{center}
\end{figure*}

Furthermore, Fig.~\ref{fig7} presents several representative failure cases of taxonomy induction using the BoostTaxo zero-shot method. Specifically, Fig.~\ref{fig7}(a) shows hierarchy flattening, where intermediate terms are omitted and multiple descendants are directly attached to a higher-level parent. Fig.~\ref{fig7}(b) illustrates excessive hierarchy depth, where unnecessary intermediate terms are introduced between terms that should be directly connected. Fig.~\ref{fig7}(c) demonstrates sibling node confusion, where a term is incorrectly attached to a sibling branch within a locally relevant region. Fig.~\ref{fig7}(d) shows cross-branch misattachment, where a term is assigned to an incorrect semantic branch rather than its correct one. These examples reveal that taxonomy induction errors arise not only from missing or redundant hierarchical levels, but also from incorrect local attachments and semantic branch misclassification.

\section{CONCLUSIONS}
Taxonomy induction is a fundamental task for organizing concepts into explicit and interpretable semantic hierarchies, yet existing methods often suffer from limited generalization, high computational cost, or insufficient structural awareness. In this paper, we proposed BoostTaxo, a boosting-style zero-shot taxonomy induction framework that performs parent identification in a coarse-to-fine manner. Specifically, the framework first enriches term semantics through retrieval-augmented definition refinement, then employs a lightweight LLM to perform efficient large-scale candidate parent filtering, and further uses a large-scale LLM to rank and score candidate parents. To improve structural reliability, we introduced a structure-aware calibration strategy that incorporates both global and local structural features to refine candidate edge weights before constructing the final taxonomy with maximum spanning arborescence.

Extensive experiments on three public benchmark datasets, namely WordNet, DBLP, and SemEval-Sci, demonstrate that the proposed method achieves strong and generally state-of-the-art performance in zero-shot taxonomy induction. In particular, BoostTaxo (GPT-4o) achieved the best Ancestor-F1 on WordNet (82.79) and SemEval-Sci (78.39), and the best Edge-F1 on all three datasets, including 63.50 on WordNet, 48.55 on DBLP, and 60.88 on SemEval-Sci. These results verify the effectiveness of the proposed coarse-to-fine framework in balancing efficiency, accuracy, and structural consistency. The ablation studies further confirm the importance of Hybrid Parent Candidate Selection and LLM-based Score Calibration with Structural Features. Hybrid candidate generation effectively narrows the search space and improves the balance between precision and recall, while structure-aware calibration further enhances the reliability of parent selection by leveraging both global and local taxonomy signals. Additional analysis also shows that the choice of Top$_{k2}$ should be adapted to taxonomy scale, with a moderate value providing a better trade-off for larger and more complex taxonomies.

Despite the promising results, the proposed method still exhibits several failure patterns, including hierarchy flattening, excessive hierarchy depth, sibling node confusion, and cross-branch misattachment, indicating that zero-shot taxonomy induction remains challenging in complex semantic structures. Future work will focus on improving robustness to these error types, exploring adaptive candidate selection and calibration strategies, and extending the framework to broader domains and larger-scale real-world taxonomy construction scenarios.




\section*{Acknowledgment}
The work is supported by TRENoP (Swedish Strategic Research Area in Transportation) and Digital Futures at KTH, Stockholm, Sweden. 

\bibliographystyle{IEEEtran}

\bibliography{Bibliography}

\begin{IEEEbiography}[{\includegraphics[width=1in,height=1.25in,clip,keepaspectratio]{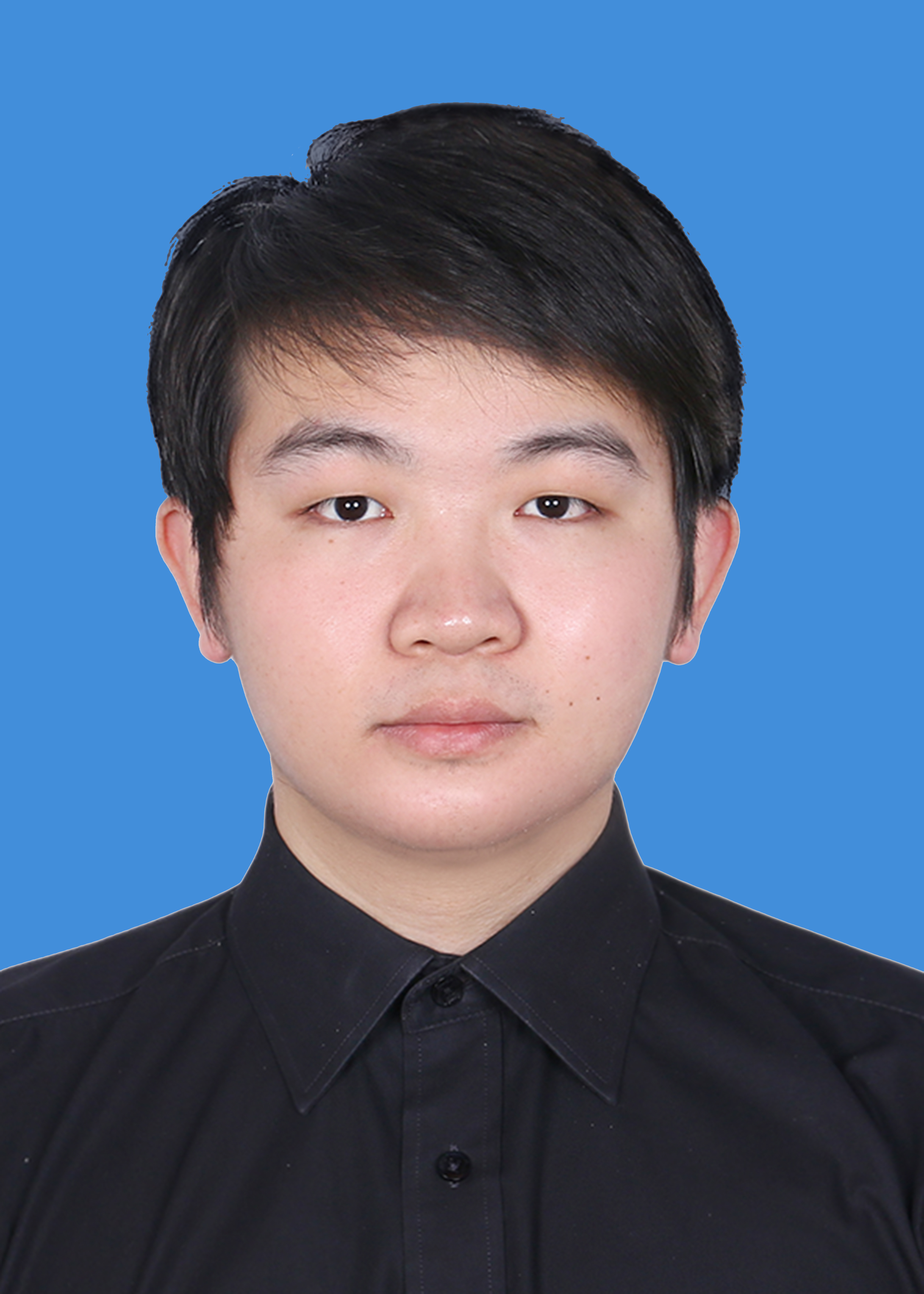}}]{Yancheng Ling} received his B.E. degrees in Traffic Engineering and Internet of Things Engineering from East China Jiaotong University in 2017, followed by his M.S. and Ph.D. degrees in Traffic Information Engineering and Control from South China University of Technology in 2019 and 2024, respectively. He was a visiting Ph.D. student at the Department of Civil and Architectural Engineering at KTH Royal Institute of Technology from 2022 to 2023. Currently, he is a postdoctoral researcher at KTH Royal Institute of Technology. His research interests primarily focus on natural language processing (NLP) and the application of large language models (LLMs) in transportation knowledge graph.
\end{IEEEbiography}

\begin{IEEEbiography}[{\includegraphics[width=1in,height=1.25in,clip,keepaspectratio]{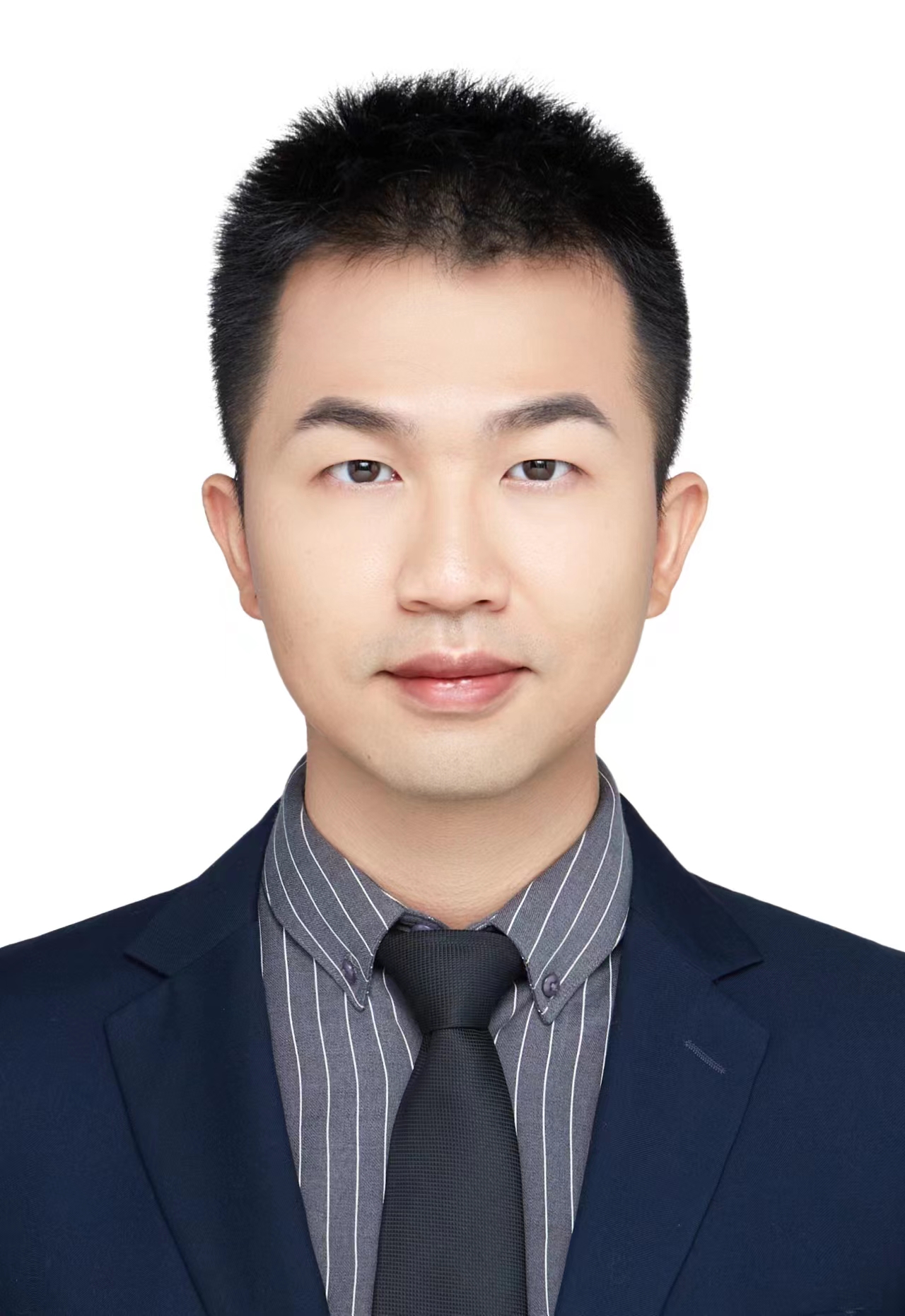}}]{Zhenlin Qin} received the B.S. degree and M.S. degree in South China University of Technology, Guangzhou, China. He is pursuing the Ph.D. degree in transport science at KTH Royal Institute of Technology. His research interest includes large language models (LLMs) and individual mobility modeling.
\end{IEEEbiography}

\begin{IEEEbiography}[{\includegraphics[width=1in,height=1.25in,clip,keepaspectratio]{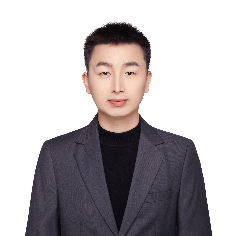}}]{Leizhen Wang} received the M.S. degree in transportation engineering from Southeast University, Nanjing, China, in 2021. He is currently working as a Ph.D. student with Department of Data Science \& Artificial Intelligence, Faculty of Information Technology, Monash University, Australia. His research interests include large language models, reinforcement learning, and intelligent transport systems.
\end{IEEEbiography}

\begin{IEEEbiography}[{\includegraphics[width=1in,height=1.25in,clip,keepaspectratio]{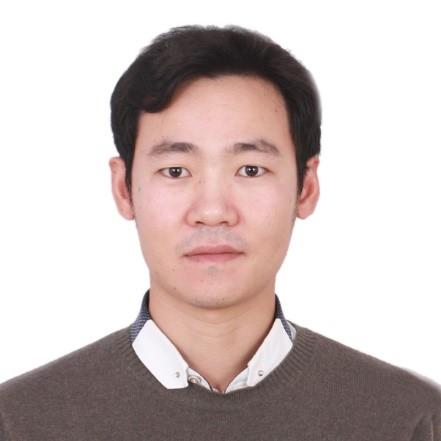}}]{Zhenliang Ma} is an Associate Professor in Road Traffic Engineering at KTH Royal Institute of Technology. His research is mainly involved in statistics, machine learning, computer science based modeling, simulation, optimization and control within the framework of selected mobility-related complex systems, which are: intelligent transport systems and multimodal mobility systems.
\end{IEEEbiography}

\vfill

\end{document}